\documentclass[10pt,twocolumn,letterpaper]{article}

\usepackage[pagenumbers]{cvpr} % To force page numbers, e.g. for an arXiv version
\usepackage{fancyhdr}
\fancypagestyle{footnotepage1}{\fancyhf{}\fancyfoot[L]{\footnotesize Accepted to Image Matching Workshop CVPR 2022. 
\\ © 2022 IEEE. Personal use of this material is permitted. Permission from IEEE must be
obtained for all other uses, in any current or future media, including
reprinting/republishing this material for advertising or promotional purposes, creating new
collective works, for resale or redistribution to servers or lists, or reuse of any copyrighted
component of this work in other works.}}

\usepackage[]{graphicx}
\usepackage{amsmath}
\usepackage{amssymb}
\usepackage{booktabs}

\usepackage[accsupp]{axessibility}  % Improves PDF readability for those with disabilities.

\usepackage[pagebackref,breaklinks,colorlinks]{hyperref}

\usepackage[capitalize]{cleveref}
\crefname{section}{Sec.}{Secs.}
\Crefname{section}{Section}{Sections}
\Crefname{table}{Table}{Tables}
\crefname{table}{Tab.}{Tabs.}

\def\cvprPaperID{?} % *** Enter the CVPR Paper ID here
\def\confName{IMW CVPR}
\def\confYear{2022}

\usepackage{preamble}

\usepackage{capt-of,etoolbox}
\makeatletter
\apptocmd\@maketitle{{\myfigure{}\par}}{}{}
\makeatother

\begin{document}

\newcommand\myfigure{%
\vspace{-.2cm}
\centering
    \includegraphics[width=0.95\columnwidth]{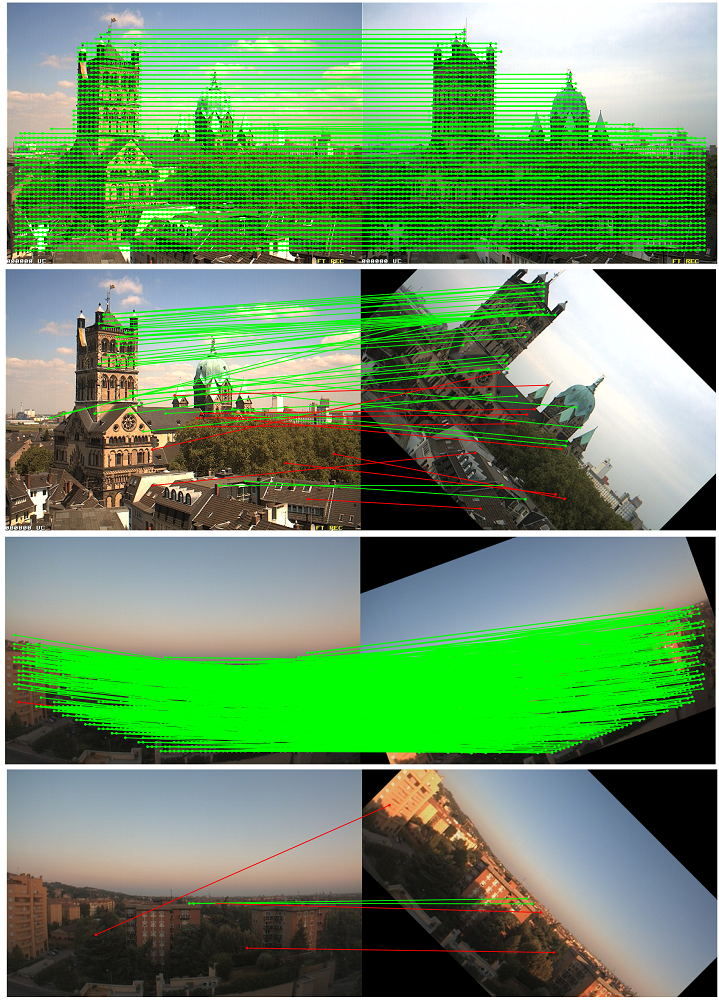}
    \hspace{.05\columnwidth}
    \includegraphics[width=0.95\columnwidth]{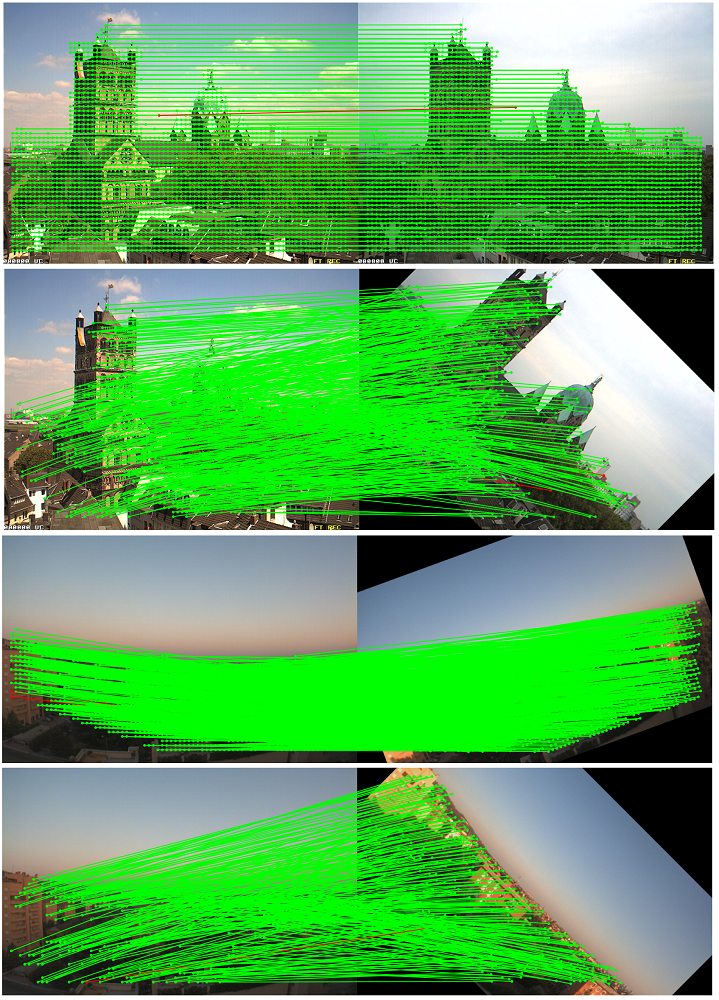}
\captionof{figure}{\emph{Left:} LoFTR \cite{sunLoFTRDetectorFreeLocal2021} finds good matches between images under illumination and small viewpoint changes, but the performance deteriorates completely under large rotation changes. \emph{Right:} Our model SE2-LoFTR-4$^\star$, which is obtained by making the LoFTR backbone CNN rotation equivariant, performs well both in the situations where the original LoFTR does, and under large rotation changes.
The image pairs are from two HPatches illumination sequences, the bottom three pairs are augmented by inplane rotations of $\ang{45}$, $\ang{20}$ and $\ang{45}$ respectively. Matches with reprojection error under $10$px are green, others red.}
\label{fig:match_comparison}
\vspace{.3cm}
}

%%%%%%%%% TITLE - PLEASE UPDATE
\title{ A case for using rotation invariant features in state of the art feature matchers}

\author{
\begin{tabular}{cc}
      \phantom{space}Georg Bökman 
     & \phantom{sp}Fredrik Kahl 
     \\
     \multicolumn{2}{c}{\texttt{\{bokman, fredrik.kahl\}@chalmers.se}}
     \\
     \multicolumn{2}{c}{Chalmers University of Technology} 
\end{tabular}
}

\maketitle

%%%%%%%%% ABSTRACT
\begin{abstract}\vspace{-.48cm}
   The aim of this paper is to demonstrate that a state of the art
feature matcher (LoFTR) can be made more robust to rotations
by simply replacing the backbone CNN with a steerable CNN which is equivariant to translations and image rotations. It is experimentally shown that this boost is obtained without reducing performance on ordinary illumination and viewpoint matching sequences.
\end{abstract}

\thispagestyle{footnotepage1} % Uncomment for footer on arxiv version.

\section{Introduction}\vspace{-.1cm}
Finding corresponding points across images is a basic building block in many computer vision tasks including structure-from-motion, motion estimation and visual localization \cite{hartleyMultipleViewGeometry2003,Toft2022TPAMI}. Hand-crafted methods have dominated the field
for a long time, but now we see major improvements with learning-based methods, making image matching work despite strong illumination changes and wide baselines.

Classical descriptors have typically been rotation invariant by design.
In the deep learning era, rotation invariant descriptors have been researched, but state of the art feature extractors usually rely on ordinary CNNs for keypoint (or dense) description, which means that the obtained descriptors are not rotation invariant.
In this paper we show that rotation invariant features can easily be obtained by using steerable CNNs for feature extraction and our experimental results indicate that while this leads to better matching of rotated images it does not deteriorate matching of non-rotated images.

We use LoFTR as a base architecture and perform extensive matching on HPatches and geometrically transformed versions of HPatches. Three different equivariant alternatives are developed and experimentally compared with the baseline.

\section{Related work and relevant theory}
This work stands on the shoulders of two giants -- the image matching literature and the literature on group equivariant neural networks.
Next, we mention related work from each of these two areas.

\subsection{Matching}
The goal of matching (sub-)pixels in two images is typically to use the obtained correspondences to estimate some form of transformation between the two images. Examples include estimating an essential matrix, a fundamental matrix or a homography \cite{hartleyMultipleViewGeometry2003}.

The classical matching pipeline can be broken up into four steps.
\begin{enumerate}
    \item  \emph{Detection.} Finding locations of interesting keypoints in both images.
    \item  \emph{Description.} Encoding the image content around these keypoints. The encodings are often called descriptors or features.
    \item  \emph{Matching.} Matching the obtained descriptors to determine which locations in the two images correspond to each other.
    \item \emph{Filtering.} Processing the matches to filter out outliers, i.e.\ incorrect matches.
\end{enumerate}

The two first steps are usually coupled, while the last steps have been more independent.
Many recent neural network based approaches however incorporate several of the steps end-to-end.
The most famous method for detection and description is SIFT \cite{loweDistinctiveImageFeatures2004}, where the detection is based on finding extrema in scale-space and the description of each keypoint is based on image gradients of close by image patches. SIFT descriptors are (approximately) rotation invariant by design.
Other classical rotation invariant descriptors include \cite{brisk, orb, surf}.
The archetypal matching method is nearest neighbour matching and the most common outlier filtering method is RANSAC \cite{ransac}.

CNN-based approaches for detection and description of keypoints include for instance D2-Net, R2D2-Net and DISK \cite{dusmanu2019cvpr, r2d2, disk}, but most related to our approach are the ones that use rotation invariant features such as LIFT, GIFT and others \cite{yiLIFTLearnedInvariant2016, liuGIFTLearningTransformationInvariant2019, lencLearningCovariantFeature2016, yiLearningAssignOrientations2016}.
While we believe that all of these approaches could be used to good effect in combination with more recent CNN based matching pipelines to obtain the benefit of rotation invariant features, none of the required modifications would be as straightforward or natural as just directly replicating a CNN architecture and replacing the layers with their steerable analogues, which is what we suggest.

Another line of research skips detection completely and instead hinges on densely describing every pixel in both images (or downsampled versions).
This approach has recently been used with great success in \cite{sunLoFTRDetectorFreeLocal2021,trune-etal-iccv-2021, edstedtDeepKernelizedDense2022}. We will base our experiments on LoFTR \cite{sunLoFTRDetectorFreeLocal2021}, which uses a CNN for dense feature description and transformer layers for further feature processing and matching. The approach is described in more detail in Section~\ref{sec:loftr}.

Learning-based methods have also been applied to the matching step. In \cite{sarlinSuperGlueLearningFeature2020}, a graph neural network is trained to perform matching, similar to our base model LoFTR.

Once a set of matches (with outliers) has been obtained, deep learning can also be used 
for the filtering step or to directly regress the transformation between the two images (e.g.\ a homography or an essential matrix).
Approaches include \cite{yiLearningFindGood2018, zhang2019oanet, sunACNeAttentiveContext2021, Zhong_2021_ICCV} and the rotation invariant \cite{bokmanZZNetUniversalRotation2021}.

\subsection{Group equivariant neural networks}
We will use the framework of $\mathrm{E}(2)$ equivariant steerable CNNs \cite{weiler_cesa_2019} to construct rotation invariant dense descriptors.
Let us first briefly describe the broader context of group equivariant neural networks before going into a bit more detail on the $\mathrm{E}(2)$-steerable ones.

A key property of CNN layers is that they are translation equivariant, i.e.\
if $T$ denotes a translation, $L$ a CNN layer and $I$ an image, then $L(T(I)) = T(L(I))$.
This property makes CNNs process image features equivalently regardless of where in an image they appear.

Extensive research has been done on describing neural networks that are equivariant under more general groups than translations \cite{cohen2016group, cohen2019gauge, kondorGeneralizationEquivarianceConvolution2018, cohenEquivariantConvolutionalNetworks2021, langWignerEckartTheoremGroup2021, weilerCoordinateIndependentConvolutional2021, aronssonHomogeneousVectorBundles2021}.
Under mild assumptions, linear layers that are group equivariant have been shown to be required to have a convolutional form.
The theory of linear equivariant layers is formulated in terms of \emph{representation theory} \cite{fulton-harris}.
% Lately, non-linear layers have however shown promise to improve on linear layers \cite{general_equiv, equiv_transformer} and such layers are an active are of research.

More broadly, group equivariant neural networks are part of the growing framework of geometric deep learning \cite{bronsteinGeometricDeepLearning2021, gerkenGeometricDeepLearning2021}.

\subsubsection{\texorpdfstring{$\mathrm{E}(2)$}{E(2)} equivariant steerable CNNs}\label{sec:steerable}
Steerable CNNs \cite{cohen-steerable, weiler3DSteerableCNNs2018, weiler_cesa_2019} are examples of group equivariant neural networks.
Let us sketch the most relevant parts of representation theory to be able to explain steerable CNNs.
In the following we will think of $G$ as being a subgroup of $\mathrm{SO}(2)$, but the theory applies in more general settings.

A tuple $(\mathbb R^m, \rho)$ is said to be a representation of $G$ if $\rho$ is a function that associates to each $g\in G$ an $m\times m$ matrix $\rho(g)$, such that $\rho(g_1g_2)=\rho(g_1)\rho(g_2)$ for all $g_1, g_2\in G$ (succinctly, $\rho$ is a group homomorphism) 
\footnote{More generally, one can instead of $\mathbb R^m$ consider an arbitrary vector space $V$ and let $\rho$ be a group homomorphism $G\to \mathrm{GL}(V)$.}.
This means that the group product in $G$ is encoded in $m\times m$ matrix multiplication, however
this encoding is not required to be \emph{faithful}, i.e. one to one.
A simple valid representation for any group and any vector space $\mathbb R^m$ is the trivial representation where for all $g$, $\rho(g)$ is the identity matrix\footnote{Technically this amounts to $m$ copies of the trivial representation, which is one dimensional.}.

In an ordinary CNN, the input image is processed into feature maps, which associate to every pixel a vector in $\mathbb R^m$ with $m$ channels.
The analogues in $G$-steerable CNNs of feature maps are called feature fields.
Like a feature map, a feature field associates each pixel to a feature vector, but whereas a CNN feature vector has no specific structure, a vector in a feature field is interpreted to lie in a representation $(\mathbb R^m, \rho)$ of the group $G$.
Concretely, this means that the feature at each pixel is a vector $v\in \mathbb R^m$ and when the pixel grid is acted on by $g\in G$ (i.e. rotated), $v$ is changed to $\rho(g)v$.
An input image corresponds to a feature field in the trivial representation -- when the image is rotated this does not change the actual RGB-values, it only changes their positions.
Features in the trivial representation are said to be $G$-\emph{invariant}.
In contrast, consider rotating an image gradient field.
The gradient vectors change position and they also rotate the same amount as the underlying pixel grid.
A gradient field is a representation $(\mathbb R^2, \rho)$ where $\rho(g)$ is the 2D rotation matrix associated with the rotation $g$.

In our experiments we use \emph{regular representations} of finite subgroups $G$ of $\mathrm{SO}(2)$.
We refer to \cite{weiler_cesa_2019, fulton-harris} for a rigorous definition of regular representations, but note that they are a way to associate a feature to each group element of $G$.
In other words, regular representations have the same number of dimensions as $G$ has elements, e.g. if $G$ has four elements then the regular representation has the vector space $\mathbb R^4$.
The matrices $\rho(g)$ in this case are permutation matrices which faithfully encode the group multiplication of $G$.
Taking multiple copies of the regular representation we can construct feature fields with any multiple of $|G|$ number of channels.

%Let us also, by way of a concrete example, mention what \emph{regular representations} are, as they are used in our experiments. Consider the finite group $C_4<\mathrm{SO}(2)$ of quarter rotations.
%$C_4$ only contains four elements and so we can design a feature field with one dimension per group element. The canonical way to do this is by using the regular representation.\footnote{For the reason why this is canonical, see for instance \cite{fulton-harris}.}
%The regular representation of $C_4$ is $(\mathbb R^4, \rho)$ where $\rho$ maps the rotation of %$n\cdot\ang{90}$ to the cyclic permutation matrix
%\[ \begin{pmatrix}
%0 & 0 & 0 & 1 \\
%1 & 0 & 0 & 0 \\
%0 & 1 & 0 & 0 \\
%0 & 0 & 1 & 0 
%\end{pmatrix}^n, \]
%for $n=0,1,2,3$.

Designing a steerable CNN architecture involves not only specifying the number of channels $m$ in each feature field, but also what representations these channels should be interpreted as belonging to.
In \cite{cohen-steerable, weiler_cesa_2019} it is shown how given a specification of input and output feature fields one can use the theory of irreducible representations \cite{fulton-harris} to solve for the linear layers/convolution kernels that are $G$-equivariant between the inputs and outputs.
This leads to even more weight sharing than using ordinary convolutions.
We refer to \cite{weiler_cesa_2019} for details. They also provide the Pytorch package \texttt{e2cnn} which we use for our experiments.

\section{Models}
The base model for our experiments is LoFTR \cite{sunLoFTRDetectorFreeLocal2021}.
To investigate the usefulness of rotation invariant features, we evaluate the original LoFTR as well as  three different versions with steerable CNNs as feature extracting backbone.

\subsection{LoFTR}\label{sec:loftr}
LoFTR (Local Feature TRansformer) is a four stage deep neural network for finding point correspondences between two images $I^A$, $I^B$ of the same scene. Briefly, the four stages are as follows. For more details we refer to \cite{sunLoFTRDetectorFreeLocal2021}.
\begin{enumerate}
    \item A feature pyramid CNN is used do extract features in the two images. This CNN is referred to as the \emph{backbone}.
    Features are obtained both at a coarse level (standard: 1/8 of the image size) and a fine level (standard: 1/2 of the image size).
    \item Positional encodings are added to the coarse features, which are then fed into a type of transformer called a \emph{LoFTR module}.
    \item The transformed coarse features are matched between the images using a differentiable matching module. As part of this matching, a confidence value is computed for each possible correspondence between the coarse feature maps. The output is a set of matching positions $\tilde i^A, \tilde i^B$ in the two images which are mutual most confident matches and whose matching confidences are above a threshold hyperparameter $\theta_c$.
    \item Around the predicted locations of the coarse matches, patches of the fine level feature maps are cropped out and processed in another LoFTR module.
    The output corresponding patches are used to compute matches from each $\tilde i^A$ to a subpixel position $\hat i^B$.
\end{enumerate}

LoFTR is trained using a loss on the coarse match confidences as well as a loss on the fine matches.
At validation/test time, after LoFTR has been used to find good matches between two images, RANSAC is used to estimate for instance an essential matrix or a homography.

The backbone CNN is a ResNet \cite{heDeepResidualLearning2015a} style feature pyramid network \cite{linFeaturePyramidNetworks2017a}. It has an initial stride 2 convolution halving the image dimensions, followed by three ResNet blocks with strides 1, 2, and 2, i.e.\ with outputs of 1/2, 1/4 and 1/8 of the original image dimensions respectively. 
This is followed by layers that iteratively upsample to 1/2 of the original image dimensions.
Coarse features are extracted at the end of the downsampling blocks and fine features at the end of the upsampling.

%\textcolor{red}{more details here?}

\begin{table}
\small
  \centering
    \begin{tabular}{lrrr}
          &       & \multicolumn{1}{l}{Relative} &  \\
          &   \multicolumn{1}{l}{Number of}     & \multicolumn{1}{l}{number of} & \multicolumn{1}{l}{Time} \\
          & \multicolumn{1}{l}{learnable} & \multicolumn{1}{l}{intermediate} & \multicolumn{1}{l}{for last} \\
          & \multicolumn{1}{l}{parameters} & \multicolumn{1}{l}{backbone} & \multicolumn{1}{l}{training} \\
          & \multicolumn{1}{l}{in backbone} & \multicolumn{1}{l}{features} & \multicolumn{1}{l}{epoch} \bigstrut[b]\\
    \hline
    LoFTR & 5.9M  & 100\% & 1h 53m \bigstrut[t]\\
    SE2-LoFTR-4$^\star$ & 1.1M  & 100\% & 1h 56m \\
    SE2-LoFTR-4 & 4.1M  & 200\% & 2h 28m \\
    SE2-LoFTR-8$^\star$ & 540k  & 100\% & 1h 58m \\
    \end{tabular}%
    \caption{Brief comparison of the different models considered.}
  \label{tab:model_comparison}%
\end{table}%

\subsection{SE2-LoFTR}
%\begin{figure*}
%    \centering
%    \begin{tikzpicture}[node distance=.7cm and .7cm]
%        \node[rotate=90] (in) {Input};
%        \node[in_conv, right=of in] (conv1) {$1 \to 128$};
%        \node[block, right=of conv1] (conv2) {$128 \to 128$};
%        \node[block_s2, right=of conv2] (conv3) {$128 \to 196$};
%        \node[block_s2, right=of conv3] (conv4) {$196 \to 256$};
%        \node[upsample_1x1, right=of conv4] (conv5) {$256 \to 196$};
%        \node[upsample_1x1, right=of conv5] (conv6) {$196 \to 128$};
%        \node[out_conv, right=of r1conv1] (r1conv2) {$196 \to 128$};
%        \node[rotate=90, rectangle, on grid, right=of fc3] (out) {Fine features};
%        \draw[bend left,->,thick] ($(r2conv3.east)+(0.35,0)$) to node [auto] {$+$} ($(fc1.east)-(0.39,0)$);
%    \end{tikzpicture}
%    \caption{The steerable CNN backbone for the SE2-LoFTR models. The \textcolor{green}{green} layers consist of $3\times3$-convolutions and the \textcolor{blue}{blue} layers of $1\times1$-convolutions. Between consecutive convolution layers, there are ReLU activations. The notation in the boxes means ``{$\text{number of input channels}\to\text{number of output channels}$}''. Skip connections are shown by the arrows marked with $+$. The \ugreendotted{dotted} layers use convolution with stride $2$, while all others use stride $1$. Every $3\times3$-convolution uses padding of one pixel around the image.}
%    \label{fig:cnn}
%\end{figure*}

We create three LoFTR variants by replacing the CNN backbone with steerable CNNs.
This modification is easily implemented with the help of the \texttt{e2cnn} Pytorch package \cite{weiler_cesa_2019}. We call the modified models SE2-LoFTR as we will consider convolutions that are steerable under certain subgroups $G$ of $\mathrm{SO}(2)$, which combined with translation equivariance means that the layers will be equivariant under subgroups of $\mathrm{SE}(2)$, the group of rotations and translations.

As explained in Section~\ref{sec:steerable}, implementing a $G$-equivariant steerable CNN involves selecting for each layer, which representation of $G$ the features should be interpreted as lying in.
In \cite{weiler_cesa_2019} the authors found that using regular representations of finite subgroups of $\mathrm{SO}(2)$ works very well compared to other choices and so we will use this approach.
%Using features in the regular representation of $G$ amounts to applying a G-CNN a la \cite{cohen2016group}, however the parametrization of the filters differs in the \texttt{e2cnn} implementation and the implementation proposed in \cite{cohen2016group}. For details, see \cite{weiler_cesa_2019}.

We will consider $G$ as being either the group $C_4$ of quarter rotations in the plane or the group $C_8$ of eighth rotations.
The features that are extracted by the backbone will lie in the trivial representation of $G$ and hence be invariant to quarter or eighth rotations respectively.
Having this type of discrete approximation of rotation invariance has been seen to work well for image classification \cite{cohen2016group, bekkers2018, weiler_cesa_2019}.

Input and output features are defined to be in the trivial representation of $G$ and all intermediate features are chosen to be in the regular representation of $G$.
However, the feature pyramid structure of the LoFTR backbone means that there are features output at the middle of the network. As we want these coarse features to be rotation invariant rather than living in the regular representation of $G$, we add a readout layer going from the middle of the network (where the features are in the regular representation) to features in the trivial representation of $G$. The outputs of this readout layer are used as coarse features.

The reason for not using trivial representations throughout the network is expressivity. Using regular representations allows for much more expressive layers \cite{weiler_cesa_2019}.

An advantage of using the regular representation rather than particular other for the features is that we can apply pointwise nonlinearities while preserving equivariance.
Thus we use an ordinary \texttt{ReLU} nonlinearity as in the original LoFTR backbone.
\texttt{BatchNorm2d} layers are replaced by their $G$-equivariant analogues \texttt{InnerBatchNorm} layers.
%These normalize over the batch, spatial and group dimensions.

The following is a list of the three SE2-LoFTR variants that we experiment with.
\begin{itemize}
    \item \emph{SE2-LoFTR-4$^\star$.} Here the LoFTR backbone is changed to be $C_4$-steerable. $C_4$ is the group of quarter rotations. For every four channels in an intermediate feature map of the LoFTR backbone we use one regular representation in the steerable feature field. Since the regular representation has $|C_4|=4$ dimensions, this makes the total number of channels equal between SE2-LoFTR-4$^\star$ and LoFTR. However, SE2-LoFTR-4$^\star$ has far fewer learnable parameters than LoFTR (see Table~\ref{tab:model_comparison}) because of the increased weight sharing in the equivariant layers.
    \item \emph{SE2-LoFTR-4.} Like SE2-LoFTR-4$^\star$ but for every \emph{two} channels in an intermediate feature map of the LoFTR backbone we use one regular representation channel.
    This brings the number of parameters closer to that of LoFTR but increases the number of features and hence both the model size and compute time.
    \item \emph{SE2-LoFTR-8$^\star$.} Like SE2-LoFTR-4$^\star$ but the considered group is $C_8$ -- the group of eighth rotations. For every eight channels in an intermediate feature map of the LoFTR backbone we use one regular representation copy in the feature field.
\end{itemize}

Indicating the models preserving the number of channels of the original model with an asterisk is consistent with \cite{weiler_cesa_2019}. In practice, these are the most similar to the original model in terms of compute and memory consumption but not in terms of number of trainable parameters.

In Table~\ref{tab:model_comparison} we list information about the four models considered in this paper. The slight overhead in training time for SE2-LoFTR-$\{\text{4,8}\}^\star$ over LoFTR is a consequence of the steerable layers, but this overhead can be alleviated at test time, by converting the steerable layers to ordinary CNN layers (see \cite[Section~2.8]{weiler_cesa_2019}).

\subsubsection{A comment on the positional encoding}
Before being passed into the LoFTR module, the coarse features are updated with positional encodings to enable the network to differentiate between absolute positions in the images.
This means that the obtained features are not actually rotation invariant, but the fact that a part of them is invariant can still facilitate matching of image patches related by a rotation.
    
\subsubsection{Implementation details}
Our code is a modification of the official LoFTR implementation and will be made available at \href{https://github.com/georg-bn/se2-loftr}{github.com/georg-bn/se2-loftr}. In particular, all hyperparameters are kept fixed from LoFTR, but we train on eight GPUs rather than sixteen, which reduces the batch size by a factor of 2. This should however not affect performance too drastically as the LoFTR implementation scales relevant hyperparameters with the batch size.
We use the \emph{dual-softmax} version of the LoFTR matching in all experiments as this version was best performing in \cite{sunLoFTRDetectorFreeLocal2021}.
During testing, we fix the matching confidence threshold $\theta_c$ to $0.2$, the same value as is used during training.
Tuning this parameter for a specific test set can improve performance.
We train all models on identical cluster compute nodes containing eight NVIDIA A100-SXM4-40GB GPUs.

\section{Datasets/benchmarks}
We train on MegaDepth \cite{megadepth}, and test on HPatches sequences \cite{balntasHPatchesBenchmarkEvaluation2017} as well as geometrically modified versions of HPatches sequences.

\subsection{MegaDepth}
We train the models on MegaDepth \cite{megadepth}, using the exact same setup as LoFTR. MegaDepth consists of 196 sets of images collected from the internet, each set corresponding to a particular tourist attraction. The task is to determine the relative pose given two images in the same set. Following LoFTR \cite{sunLoFTRDetectorFreeLocal2021}, validation is done on sequences 0015 and 0022\footnote{0015 is St. Peter's Square and 0022 is Brandenburger Tor. Confusion has previously arised from the fact that Sacre Coeur was labelled 0022 in the DISK-paper \cite{disk}, leading to LoFTR claiming to evaluate on Sacre Coeur, while they do evaluate on Brandenburger Tor.}. Training is done on the same subset of scenes as in LoFTR. There is no separate test set and early stopping is performed on the validation set. We hence argue that the results on MegaDepth should be interpreted as validation results only.

\subsubsection{Metrics}
The performance on MegaDepth is measured as in \cite{sarlinSuperGlueLearningFeature2020, sunLoFTRDetectorFreeLocal2021} in terms of the area under curve (AUC) of the pose accuracy up to a specific threshold.
The considered thresholds are $\ang{5}, \ang{10}$ and $\ang{20}$.
The pose accuracy at angle $a$ is defined as the proportion of relative poses whose rotation and translation are both within $a$ from the ground truth.
Each AUC is normalized so that the max score is $100\%$.
Early validation stopping is done on the AUC@$\ang{10}$.

\begin{figure}
    \centering
    \includegraphics[width=0.37\columnwidth]{figures/3.png}
    \hspace{0.3cm}
    \includegraphics[width=0.37\columnwidth]{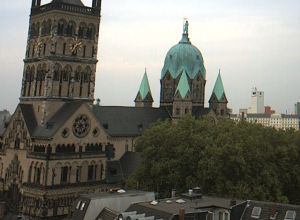}
    \caption{Left: An image from the HPatches sequences. Right: The modified image in HPatches-h0.3.}
    \label{fig:hpatch_warp}
\end{figure}

\begin{table}
    \centering
    \small
\begin{tabular}{lccc}\toprule
            \multicolumn{1}{r}{Pose error AUC (\%)} &   @5° &  @10° & @20° \\
            \midrule 
            LoFTR (paper) & {\cellcolor{yellow}}52.8 & {\cellcolor{yellow}}69.2 & 81.3 \\
            LoFTR (re-trained) & 52.2 & 68.9 & 81.0\\
            SE2-LoFTR-4$^\star$ & 51.3 & 68.0 & 80.3\\
            SE2-LoFTR-4 & 52.6 & {\cellcolor{yellow}}69.2 & {\cellcolor{yellow}}81.4 \\ 
            SE2-LoFTR-8$^\star$ & 50.1 & 66.9 & 79.4
            \end{tabular}
    \caption{Validation results on MegaDepth.}
    \label{tab:md}
\end{table}

\subsection{HPatches}
Testing of the models is done on HPatches sequences \cite{balntasHPatchesBenchmarkEvaluation2017} where the task is to find a homography relating two images. Again we follow the LoFTR setup. We use the original HPatches sequences and the later added ones from \cite{AMOS,DTU,HAN,OXF,PHOS,ASIFT}.
Following D2-Net \cite{dusmanu2019cvpr} eight image sequences are removed\footnote{High resolution images are removed. This removal can be considered a historical artefact stemming from the fact that some methods were not able to handle the high resolutions.} which leaves 108 sequences.
Out of these, 52 sequences contain \emph{illumination} changes and 56 contain \emph{viewpoint} changes.
Each sequence contains six images and ground truth homographies relating the first to each of the following five. We refer to the first image in each sequence as the $A$-image and to the five images in each sequence that are to be related to $A$ as the $B$-images.

We follow \cite{detoneSuperPointSelfSupervisedInterest2018} in resizing all images to size $640\times 480$ (or $480\times 640$ for upright images). This might slightly deviate from the approach used by the LoFTR authors who state ``All images are resized with shorter dimensions equal to 480.''. As the aim of the resizing is to make the AUC of corner errors comparable across images, we believe it to be reasonable to use a uniform size for all images.

\begin{table*}
\centering
\footnotesize
\begin{tabular}{lccccccccc}\toprule
            {} & \multicolumn{3}{l}{All} & \multicolumn{3}{l}{Illumination} & \multicolumn{3}{l}{Viewpoint} \\
            \multicolumn{1}{r}{MMA} &   @3px &  @5px & @10px &            @3px &            @5px &           @10px &            @3px &            @5px &           @10px \\
            \midrule 
            \multicolumn{10}{l}{\textbf{HPatches}} \\
            %LoFTR (paper) & \phantom{00.0} & \phantom{00.0} & \phantom{00.0} & \phantom{00.0} & \phantom{00.0} & \phantom{00.0} & \phantom{00.0} & \phantom{00.0} & \phantom{00.0} \\
            LoFTR (re-evaluated) & 93.4 & 95.2 & 96.1 & 98.4 & 98.7 & 99.3 & 88.9 & 91.9 & 93.1 \\LoFTR (re-trained) & 93.4 & 95.1 & 96.0 & {\cellcolor{yellow}}98.6 & {\cellcolor{yellow}}98.9 & 99.5 & 88.7 & 91.5 & 92.7 \\SE2-LoFTR-4$^\star$ & 94.3 & 96.2 & 97.3 & 98.3 & 98.7 & 99.4 & 90.5 & 93.9 & 95.3 \\SE2-LoFTR-4 & {\cellcolor{yellow}}94.5 & {\cellcolor{yellow}}96.5 & {\cellcolor{yellow}}97.5 & 98.5 & {\cellcolor{yellow}}98.9 & {\cellcolor{yellow}}99.6 & {\cellcolor{yellow}}90.8 & {\cellcolor{yellow}}94.2 & {\cellcolor{yellow}}95.5 \\SE2-LoFTR-8$^\star$ & 94.1 & 96.1 & 97.3 & 98.2 & 98.7 & 99.4 & 90.3 & 93.8 & 95.3 \\
            \midrule 
            \multicolumn{10}{l}{\textbf{HPatches-r20}} \\
            LoFTR (re-evaluated) & 86.0 & 90.5 & 92.6 & 89.5 & 94.6 & 97.1 & 82.7 & 86.7 & 88.5 \\LoFTR (re-trained) & 86.6 & 91.0 & 92.9 & 90.2 & 95.2 & 97.4 & 83.2 & 87.1 & 88.7 \\SE2-LoFTR-4$^\star$ & 89.4 & 93.8 & 95.8 & 90.7 & 95.2 & 97.4 & 88.1 & 92.5 & {\cellcolor{yellow}}94.3 \\SE2-LoFTR-4 & {\cellcolor{yellow}}89.7 & {\cellcolor{yellow}}94.0 & {\cellcolor{yellow}}95.9 & {\cellcolor{yellow}}91.1 & {\cellcolor{yellow}}95.5 & {\cellcolor{yellow}}97.6 & 88.4 & 92.5 & {\cellcolor{yellow}}94.3 \\SE2-LoFTR-8$^\star$ & 89.5 & 93.9 & 95.8 & 90.5 & 95.2 & 97.5 & {\cellcolor{yellow}}88.5 & {\cellcolor{yellow}}92.7 & {\cellcolor{yellow}}94.3 \\
            \midrule 
            \multicolumn{10}{l}{\textbf{HPatches-r45}} \\
            LoFTR (re-evaluated) & 35.0 & 41.5 & 46.2 & 37.7 & 45.2 & 50.3 & 32.5 & 38.1 & 42.4 \\LoFTR (re-trained) & 30.7 & 38.1 & 44.3 & 30.3 & 39.3 & 46.9 & 31.1 & 37.1 & 41.9 \\SE2-LoFTR-4$^\star$ & 72.6 & 83.9 & 89.2 & 69.6 & 81.9 & 88.3 & 75.3 & 85.7 & 90.0 \\SE2-LoFTR-4 & 70.9 & 84.1 & 89.9 & 68.4 & 82.7 & 89.4 & 73.3 & 85.4 & 90.4 \\SE2-LoFTR-8$^\star$ & {\cellcolor{yellow}}80.4 & {\cellcolor{yellow}}89.3 & {\cellcolor{yellow}}93.0 & {\cellcolor{yellow}}78.8 & {\cellcolor{yellow}}88.7 & {\cellcolor{yellow}}93.1 & {\cellcolor{yellow}}81.9 & {\cellcolor{yellow}}89.9 & {\cellcolor{yellow}}92.8 \\
            \midrule 
            \multicolumn{10}{l}{\textbf{HPatches-h0.3}} \\
            LoFTR (re-evaluated) & 87.7 & 92.7 & 95.0 & 88.9 & 94.6 & 97.3 & 86.5 & 90.9 & 92.9 \\LoFTR (re-trained) & 88.0 & 93.0 & 95.2 & {\cellcolor{yellow}}89.8 & {\cellcolor{yellow}}95.3 & {\cellcolor{yellow}}97.8 & 86.4 & 90.8 & 92.7 \\SE2-LoFTR-4$^\star$ & 88.9 & 94.2 & {\cellcolor{yellow}}96.6 & 89.3 & 95.0 & 97.6 & 88.6 & 93.4 & 95.6 \\SE2-LoFTR-4 & {\cellcolor{yellow}}89.3 & {\cellcolor{yellow}}94.4 & {\cellcolor{yellow}}96.6 & 89.5 & 95.0 & 97.5 & {\cellcolor{yellow}}89.1 & {\cellcolor{yellow}}93.8 & {\cellcolor{yellow}}95.8 \\SE2-LoFTR-8$^\star$ & 88.2 & 93.6 & 96.2 & 87.7 & 93.8 & 96.8 & 88.6 & 93.4 & 95.7 \
            \end{tabular}
\caption{Mean match accuracy in percent on HPatches and modified versions. HPatches-r$a$ contains images that are rotated by $a$ degrees from the original HPatches sequences and HPatches-h0.3 contains images warped using the scheme described in Section~\ref{sec:hpatch_mod}.}
\label{tab:MMA}
\end{table*}

\subsubsection{Modifications of HPatches sequences}\label{sec:hpatch_mod}
As we want to test the robustness of the models under rotations, we introduce modifications of the HPatches sequences as follows. For an angle $a\in(\ang{0}, \ang{90}]$, we rotate each $B$-image by $a$ either clockwise or anticlockwise (this is chosen randomly).
We call the obtained modified HPatches sequence HPatches-r$a$, and will consider HPatches-r20 and HPatches-r45. Examples of these sequences can be seen in Figure~\ref{fig:match_comparison}.

To test if enforcing increased robustness under rotations influences robustness under other types of geometric transformations, we introduce a further modified HPatches variant.
Each $B$-image is warped as follows.
Let the image dimensions be $h\times w$, and choose a value $s>0$ to control the amount of warping.
\begin{enumerate}
    \item For each of the four image corners, sample an offset of at most $(s\cdot h, s\cdot w)$.
    \item Warp the corners outwards from the image center by the respective offsets. I.e.\ so that the upper left corner $(0, 0)$ is moved to somewhere in the rectangle \mbox{$[-s\cdot h, 0]\times[-s\cdot w, 0]$} etc.
\end{enumerate}
This warp is determined by four points and hence corresponds to a homography, which is used to alter the ground truth homography from $A$ to $B$. 
The resulting $B$-images are slightly skewed zoom-ins of the originals.
We denote the obtained modified sequence by HPatches-h$s$, and will consider HPatches-h0.3 in the experiments.
An example image of HPatches-h0.3 is shown in Figure~\ref{fig:hpatch_warp}.

The modifications in HPatches-r$a$ and HPatches-h$s$ are carried out on top of the illumination/viewpoint transformations that are already present in HPatches. Hence, these datasets are quite a bit more challenging than the original HPatches.

\begin{table*}
\centering
\footnotesize
\begin{tabular}{lccccccccc}\toprule
            {} & \multicolumn{3}{l}{All} & \multicolumn{3}{l}{Illumination} & \multicolumn{3}{l}{Viewpoint} \\
            \multicolumn{1}{r}{Corner error AUC} &   @3px &  @5px & @10px &            @3px &            @5px &           @10px &            @3px &            @5px &           @10px \\
            \midrule 
            \multicolumn{10}{l}{\textbf{HPatches}} \\
            LoFTR (paper) & 65.9 & 75.6 & 84.6 & \phantom{00.0} & \phantom{00.0} & \phantom{00.0} & \phantom{00.0} & \phantom{00.0} & \phantom{00.0} \\LoFTR (re-evaluated) & 65.3 & 74.8 & 84.4 & 81.8 & 88.7 & 94.2 & 50.2 & 62.1 & 75.5 \\LoFTR (re-trained) & 65.3 & 75.5 & 84.4 & 81.5 & 88.4 & 94.0 & 50.5 & 63.6 & 75.6 \\SE2-LoFTR-4$^\star$ & 65.2 & 75.1 & 84.7 & 80.8 & 87.9 & 93.8 & 51.0 & 63.3 & 76.3 \\SE2-LoFTR-4 & {\cellcolor{yellow}}66.2 & {\cellcolor{yellow}}76.6 & {\cellcolor{yellow}}86.0 & {\cellcolor{yellow}}82.2 & {\cellcolor{yellow}}89.2 & {\cellcolor{yellow}}94.4 & 51.5 & 65.0 & {\cellcolor{yellow}}78.3 \\SE2-LoFTR-8$^\star$ & 65.8 & 76.0 & 85.5 & 80.4 & 87.8 & 93.8 & {\cellcolor{yellow}}52.3 & {\cellcolor{yellow}}65.2 & 77.9 \\
            \midrule 
            \multicolumn{10}{l}{\textbf{HPatches-r20}} \\
            LoFTR (re-evaluated) & 52.5 & 65.3 & 78.0 & 61.0 & 74.7 & 86.7 & 44.7 & 56.7 & 70.0 \\LoFTR (re-trained) & 51.3 & 64.7 & 77.4 & 62.8 & 76.4 & 87.6 & 40.9 & 53.9 & 68.1 \\SE2-LoFTR-4$^\star$ & 55.3 & 68.4 & 80.8 & 64.5 & 76.9 & 87.7 & 46.9 & 60.6 & 74.5 \\SE2-LoFTR-4 & 54.9 & 68.3 & 80.9 & 64.0 & 76.4 & 87.7 & 46.6 & {\cellcolor{yellow}}61.1 & {\cellcolor{yellow}}74.7 \\SE2-LoFTR-8$^\star$ & {\cellcolor{yellow}}56.1 & {\cellcolor{yellow}}68.7 & {\cellcolor{yellow}}81.1 & {\cellcolor{yellow}}65.6 & {\cellcolor{yellow}}77.8 & {\cellcolor{yellow}}88.3 & {\cellcolor{yellow}}47.5 & 60.3 & 74.6 \\
            \midrule 
            \multicolumn{10}{l}{\textbf{HPatches-r45}} \\ LoFTR (re-evaluated) & 15.1 & 22.9 & 32.0 & 16.2 & 26.5 & 37.8 & 14.2 & 19.7 & 26.8 \\LoFTR (re-trained) & 10.3 & 16.9 & 27.3 & 9.64 & 18.0 & 31.0 & 11.1 & 16.1 & 24.1 \\SE2-LoFTR-4$^\star$ & 31.0 & 47.1 & 66.3 & 31.6 & 50.3 & 70.3 & 30.5 & 44.4 & 62.8 \\SE2-LoFTR-4 & 25.4 & 42.3 & 64.0 & 24.2 & 42.9 & 66.5 & 26.7 & 41.8 & 61.8 \\SE2-LoFTR-8$^\star$ & {\cellcolor{yellow}}38.8 & {\cellcolor{yellow}}55.2 & {\cellcolor{yellow}}73.3 & {\cellcolor{yellow}}42.0 & {\cellcolor{yellow}}59.6 & {\cellcolor{yellow}}77.8 & {\cellcolor{yellow}}36.0 & {\cellcolor{yellow}}51.3 & {\cellcolor{yellow}}69.2 \\
            \midrule 
            \multicolumn{10}{l}{\textbf{HPatches-h0.3}} \\LoFTR (re-evaluated) & 44.9 & 59.0 & 74.9 & 54.4 & 69.4 & 83.7 & 36.3 & 49.6 & 66.9 \\LoFTR (re-trained) & 44.7 & 58.7 & 74.2 & 54.2 & 68.8 & 83.0 & 36.1 & 49.5 & 66.3 \\SE2-LoFTR-4$^\star$ & 45.4 & 60.0 & 75.8 & 55.5 & {\cellcolor{yellow}}70.6 & 84.0 & 36.2 & 50.3 & 68.3 \\SE2-LoFTR-4 & {\cellcolor{yellow}}45.9 & {\cellcolor{yellow}}60.5 & {\cellcolor{yellow}}76.2 & {\cellcolor{yellow}}55.7 & 70.5 & {\cellcolor{yellow}}84.2 & 37.0 & {\cellcolor{yellow}}51.4 & {\cellcolor{yellow}}69.0 \\SE2-LoFTR-8$^\star$ & 45.7 & 60.0 & 75.6 & 54.9 & 69.4 & 83.4 & {\cellcolor{yellow}}37.4 & {\cellcolor{yellow}}51.4 & 68.4 \\
            \end{tabular}
\caption{AUC of corner error in percent on HPatches and modified versions. HPatches-r$a$ contains images that are rotated by $a$ degrees from the original HPatches sequences and HPatches-h0.3 contains images warped using the scheme described in Section~\ref{sec:hpatch_mod}.}
\label{tab:HCEAUC}
\end{table*}

\subsubsection{Metrics}\label{sec:hpatch_metrics}
The AUC@x metric is used as in \cite{sunLoFTRDetectorFreeLocal2021}.
For an image correspondence $A, B$ with ground truth homography $H$, the \emph{corner error} is defined as the mean of the distances between the corners of $A$ warped by $H$ and the corners of $A$ warped by an estimated homography $\tilde H$.
The metric used is then the area under curve (AUC) of the corner error accuracy up to thresholds $3$px, $5$px and $10$px.
Each AUC score is normalized so that the max is $100\%$.

We furthermore report the mean match accuracy (MMA) at the same thresholds.
The MMA is defined as the mean proportion of matches that are within a certain threshold in reprojection error. It was used in e.g. \cite{d2-net, zhouPatch2PixEpipolarGuidedPixelLevel2021,edstedtDeepKernelizedDense2022}.

To compute these metrics, we use the \texttt{immatch} package \cite{immatch-package}.
The estimated homography $\tilde H$ is obtained by running  the OpenCV function \texttt{findHomography} on the top 1000 most confident matches with RANSAC as the outlier filtering method.
We use the default OpenCV hyperparameters for RANSAC.

The metrics are sensitive to for instance image size, RANSAC hyperparameters as well as the matching threshold $\theta_c$. As we do not know the exact settings used for the results in the LoFTR paper, the results are not directly comparable.

\section{Results}
For comparison, we retrain a LoFTR version using our training setup (in particular lower batch size) and evaluate it along with the pretrained version supplied by the authors.
The retrained version uses updated code which fixes a bug in the positional encoding.
We furthermore report the numbers from the LoFTR paper where available.

\subsection{MegaDepth}
Validation results on MegaDepth are shown in Table~\ref{tab:md}.
We see that performance correlates quite well with the number of parameters in each model, but the performance is similar across all models.
All models perform better than DRC-Net \cite{DRC} and SuperPoint+SuperGlue \cite{detoneSuperPointSelfSupervisedInterest2018, sarlinSuperGlueLearningFeature2020} which are the comparisons in the LoFTR paper \cite[Table~3]{sunLoFTRDetectorFreeLocal2021}.

\subsection{HPatches}
Qualitative results are shown in Figure~\ref{fig:match_comparison}.
Tables~\ref{tab:MMA} and \ref{tab:HCEAUC} show the test results on the ordinary HPatches sequences as well as modified versions described in Section~\ref{sec:hpatch_mod}.
The most important score is the corner error AUC in Table~\ref{tab:HCEAUC} as the MMA (Table~\ref{tab:MMA}) can be unreliable when the amount of extracted matches fluctuates. 
The difference between the LoFTR (paper) row and the LoFTR (re-evaluated) row is due to differences in how the evaluation is performed as described in Section~\ref{sec:hpatch_metrics}.

Let us point out a couple of interesting results.
\begin{itemize}
    \item The SE2 versions clearly outperform the baseline on the rotated HPatches-r datasets.
    \item The SE2 versions perform very well on the unmodified HPatches. This means that using rotation invariant features is not detrimental to general performance.
    In particular, the SE2 versions quite consistently outperform the baseline on the sequences of HPatches with viewpoint changes.
    \item The SE2 versions outperform the baseline on the warped HPatches-h0.3, showing that rotation invariant features can improve robustness under more general geometric transformations than rotations.
    \item The SE2-LoFTR-8$^\star$ model outperforms the other SE2 versions on the rotated HPatches-r datasets. Its features are by design invariant to $\ang{45}$ rotations so the great performance on HPatches-r45 is unsurprising.
\end{itemize}

\begin{table*}
\centering
\footnotesize
\begin{tabular}{lllllll}
\toprule
{} & \multicolumn{3}{l}{Corner error AUC}  & \multicolumn{3}{l}{MMA}\\
{} &                                            @3px &                                            @5px &                                           @10px &                                              @3px &                                            @5px &                                           @10px\\
\midrule
LoFTR (paper)        &  {(65.9, \phantom{00.0}, \phantom{00.0})} &  {(75.6, \phantom{00.0}, \phantom{00.0})} &  {(84.6, \phantom{00.0}, \phantom{00.0})} & & & \\
LoFTR (re-evaluated) &  {(65.3, \phantom{00.0}, \phantom{00.0})} &  {(74.8, \phantom{00.0}, \phantom{00.0})} &  {(84.4, \phantom{00.0}, \phantom{00.0})} & {(93.4, \phantom{00.0}, \phantom{00.0})} &     {(95.2, \phantom{00.0}, \phantom{00.0})} &     {(96.1, \phantom{00.0}, \phantom{00.0})} \\
LoFTR (re-trained)   &                      {(65.3, 65.8, 64.2)} &                      {(75.5, 75.5, 74.1)} &                      {(84.4, 84.6, 83.5)} & {(93.4, 93.8, 93.3)} &                         {(95.1, 95.4, 95.0)} &                         {(96.0, 96.3, 95.9)} \\
SE2-LoFTR-4$^\star$          &                      {(65.2, 65.6, 65.1)} &                      {(75.1, 75.7, 75.0)} &                      {(84.7, 85.5, 84.7)} & {(94.3, 94.5, 94.1)} &                         {(96.2, 96.4, 96.0)} &                         {(97.3, 97.5, 97.0)} \\
SE2-LoFTR-4      &                      {(66.2, 66.3, 65.5)} &                      {(76.6, 76.6, 76.0)} &                      {(86.0, 86.0, 85.6)} & {(94.5, 94.7, 94.2)} &                         {(96.5, 96.7, 96.2)} &                         {(97.5, 97.6, 97.2)} \\
SE2-LoFTR-8$^\star$          &                      {(65.8, 65.8, 64.6)} &                      {(76.0, 76.0, 74.6)} &                      {(85.5, 85.5, 84.6)} & {(94.1, 94.1, 94.0)} &                         {(96.1, 96.2, 96.0)} &                         {(97.3, 97.4, 97.2)} \\

\bottomrule
\end{tabular}

\caption{Variability among test scores of the top five validation checkpoints of a training run. The scores presented are over all ordinary HPatches sequences in the format (\mbox{\texttt{Score of checkpoint with best MegaDepth validation score}}, \mbox{\texttt{Highest score on HPatches among the five checkpoints}}, \\\mbox{\texttt{Lowest score on HPatches among the five checkpoints}}). I.e.\ the first value of each triple corresponds to the value presented in all other tables.}
\label{tab:triple}
\end{table*}

\subsection{Limitations}
We have only trained each model once due to computational limitations.
This is similar to most papers in the field.
However, we have saved checkpoints from 5 epochs of the training run, corresponding to the top 5 validation scores on AUC@$\ang{10}$ on MegaDepth.
We can evaluate each of these on HPatches to get a sense of the variability of the test scores.
Table~\ref{tab:triple} shows that the variability is not too high.

We tested implementing a naive data augmentation scheme where images were rotated by some multiple of quarter rotations, but could not get reasonable results as the coarse matching too often failed to produce any matches at all during training.
Any form of data augmentation technique would however likely require much longer training times as the training data set implicitly is enlarged.

The \texttt{e2cnn} package enables many more general representations for the feature fields of the backbone CNN. We have limited the scope of this paper to regular representations but cannot claim to know that this is the best choice.

\section{Conclusion and future work}
We argue that steerable CNNs should be a more common tool in feature matching pipelines than they currently are and have showed promising results with a very simple modification of the LoFTR model. Robustness to rotations is achieved while keeping the matching performance for non-rotated image pairs and with little computational overhead.

We note that the whole pipeline is not equivariant due to the transformer layers.
It would be interesting to investigate whether further boosts can be obtained by designing the transformer layers in an equivariant manner as well. 
%Group equivariant transformer like architectures have been discussed in for instance \cite{romero_transformer, efficient_equiv_net}.

Finally, we mention that rotation invariant features can be of great value in domains such as matching aerial images where there is no canonical image orientation or non-rigid matching where the orientations of the features might change between the images. Steerable CNN feature extraction could also be combined with the approach proposed in \cite{toft2020}, where features are obtained on rectified planar surfaces of the input images -- these rectified planar regions can have arbitrary relative planar rotations between the two images.

\nocite{olsson-etal-cvpr-2006,kahl-etal-cvpr-2001}

\section*{Acknowledgements} We thank the anonymous reviewers whose comments helped improve the paper. This work was supported by the Swedish Foundation for Strategic Research and the Wallenberg AI, Autonomous Systems and Software Program (WASP) funded by the Knut and Alice Wallenberg Foundation (KAW). The computations were enabled by the supercomputing resource Berzelius provided by the National Supercomputer Centre (NSC), funded by Linköping University and KAW.

%%%%%%%%% REFERENCES
%\newpage
{\small
\bibliographystyle{ieee_fullname}
\bibliography{egbib}
}

%%%%%%%%%% Supplementary

%\newpage
%\onecolumn
%\input{supplementary}

\end{document}